\documentclass[lettersize,journal]{IEEEtran}
\usepackage{amsmath,amsfonts}
\usepackage{algorithmic}
\usepackage{algorithm}
\usepackage{array}
\usepackage[caption=false,font=normalsize,labelfont=sf,textfont=sf]{subfig}
\usepackage{textcomp}
\usepackage{stfloats}
\usepackage{url}
\usepackage{verbatim}
\usepackage{graphicx}
\usepackage{cite}
\usepackage[verbose]{placeins}
\usepackage{amssymb}  
\usepackage{siunitx}
\usepackage{mathtools}
\usepackage{booktabs}
\usepackage{multirow}
\usepackage[table]{xcolor}  
\usepackage{booktabs}       
\usepackage{pgf}            

\usepackage{subcaption} 
\usepackage{tikz}

\usepackage{array,multirow}
\usepackage{makecell}
\usepackage[T1]{fontenc}
\usepackage{microtype}
\usepackage{tikz}
\usetikzlibrary{positioning}
\usepackage{pgfplots}
\usepackage{xspace}
\usepackage[capitalise]{cleveref}
\crefformat{equation}{Eq.(#2\color{red}#1#3)}
\crefformat{figure}{Fig.(#2\color{red}#1#3)}
\crefformat{algorithm}{Alg.(#2\color{red}#1#3)}
\definecolor{YellowGreen}{RGB}{245,250,200}
\definecolor{LightGreen}{RGB}{185,235,160}

\newcommand{\formattedparagraph}[1]{\noindent \textbf{#1}}

\begin{document}

\title{Next-Future: Sample-Efficient Policy Learning for Robotic-Arm Tasks}

\author{Fikrican \"Ozg\"ur, René Zurbr\"ugg, Suryansh Kumar$^\dagger$
\thanks{Fikrican \"Ozg\"ur was with ETH Z\"urich Switzerland.}
\thanks{René Zurbr\"ugg is with RSL Group at ETH Z\"urich Switzerland.}
\thanks{Suryansh Kumar is with Visual Computing and Computational Media Section, College of PVFA, Department of Electric and Computer Engineering, and Department of Computer Science and Engineering at Texas A\&M University, College Station, Texas, USA.}
\thanks{$\dagger$ Corresponding Author: Suryansh Kumar (Email: k.sur46@gmail.com)}
}



\maketitle

\begin{abstract}
Hindsight Experience Replay (HER) is widely regarded as the state-of-the-art algorithm for achieving sample-efficient multi-goal reinforcement learning (RL) in robotic manipulation tasks with binary rewards. HER facilitates learning from failed attempts by replaying trajectories with redefined goals. However, it relies on a heuristic-based replay method that lacks a principled framework. To address this limitation, we introduce a novel replay strategy, ``Next-Future'', which focuses on rewarding single-step transitions. This approach significantly enhances sample efficiency and accuracy in learning multi-goal Markov decision processes (MDPs), particularly under stringent accuracy requirements—a critical aspect for performing complex and precise robotic-arm tasks. We demonstrate the efficacy of our method by highlighting how single-step learning enables improved value approximation within the multi-goal RL framework. The performance of the proposed replay strategy is evaluated across eight challenging robotic manipulation tasks, using ten random seeds for training. Our results indicate substantial improvements in sample efficiency for seven out of eight tasks and higher success rates in six tasks. Furthermore, real-world experiments validate the practical feasibility of the learned policies, demonstrating the potential of ``Next-Future'' in solving complex robotic-arm tasks.
\end{abstract}


\section{INTRODUCTION}
\IEEEPARstart{E}{veryday} tasks such as object grasping and flipping, which are effortless for humans, remain a challenge in robotics. Traditional robotic-arm autonomy system or algorithmic pipelines, characterized by loosely coupled modules for perception, planning, and control, are susceptible to failure due to error propagation across these components. Consequently, robotic manipulators are predominantly deployed in structured environments, where they perform repetitive tasks through meticulously designed motion sequences, thus significantly limiting their operational scope for other tasks \cite{murray2017mathematical}.

An alternative approach involves training the entire pipeline of different robotic-arm task using a deep neural network in a end-to-end way using deep reinforcement learning (DRL). To this end, it is observed that DRL offers a novel perspective and several advantages  over traditional approaches \cite{levine2016end}. Notably, DRL approaches eliminate the necessity of manually designing intermediate representations to transition between pipeline modules. Instead, these representations are autonomously learned through the optimization process, leveraging the available data. This approach enhances the robustness of the robotic-arm functional system pipeline, effectively addressing the inherent fragility of traditional robotic control systems.

Despite its advantages, deep reinforcement learning (DRL) faces significant limitations, particularly due to its low sample efficiency \cite{yu2020meta, henderson2018deep, yu2018towards}. Such a limitation is especially problematic in robotic-arm automation tasks, where collecting experiences can be both costly and time-intensive. Additionally, model-based reinforcement learning (RL) for solving robotic-arm tasks presents two further challenges. First, robotic-arm automation are often expected to be versatile, capable of solving multiple tasks within their environment. However, the model-based RL framework typically supports only a single, predefined task, making it ill-suited for such versatility. Second, model-based RL requires the well-crafted design of reward functions tailored to specific tasks, which demands substantial domain-specific knowledge. This process becomes particularly challenging in high-dimensional tasks, where numerous factors influence the robot's performance, further complicating reward formulation.

\begin{figure}[t]
    \centering
    \includegraphics[scale=0.44]{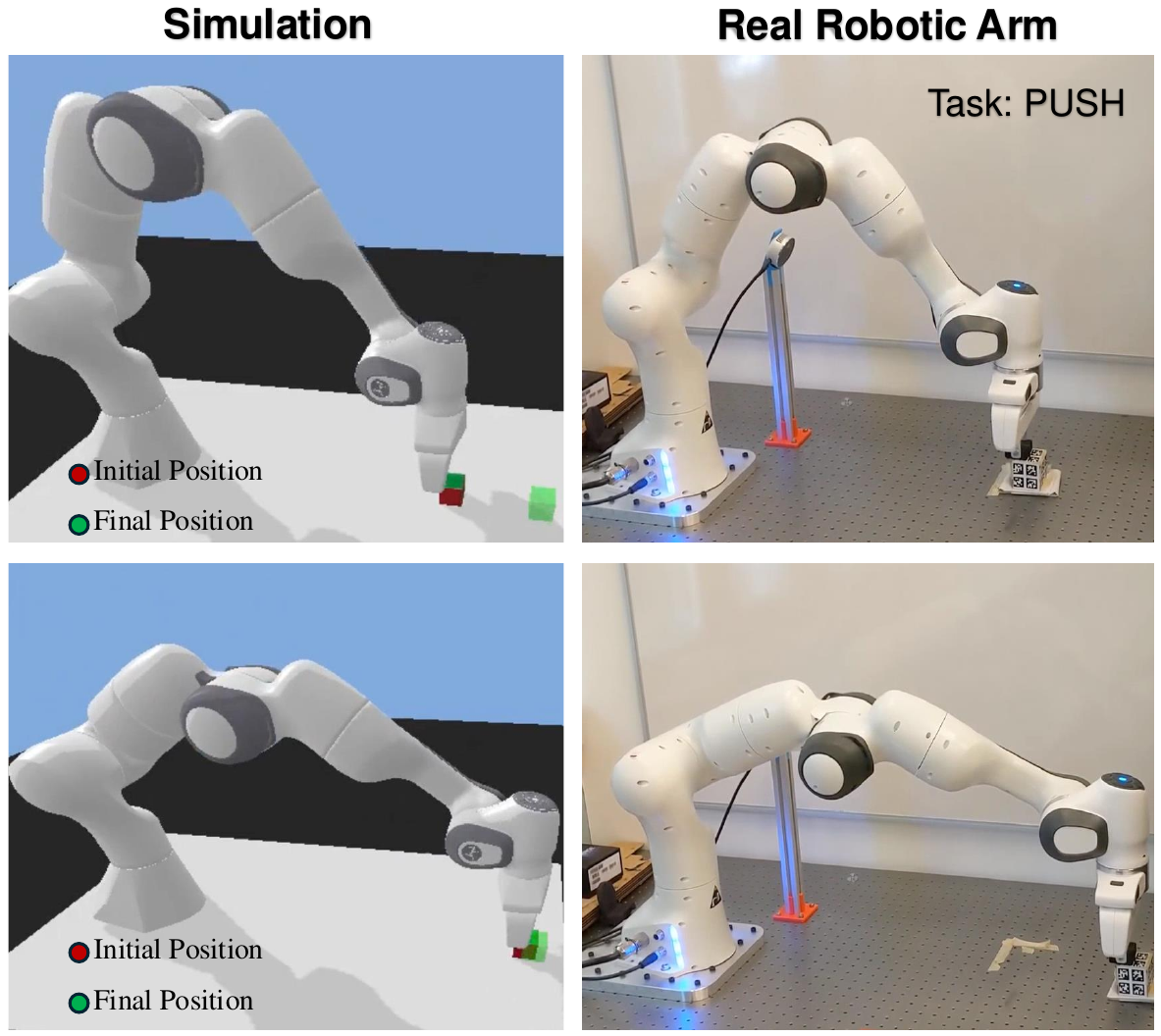}
    \caption{The robotic arm perform the task of push a cube from an initial position to specified final position. \textbf{Left column:} shows simulation environment. \textbf{Right column:} shows corresponding output on the real robotic arm platform.}
    \label{fig:teaser-function}
\end{figure}

By effectively addressing the challenges outlined above, Hindsight Experience Replay (HER) \cite{andrychowicz2017hindsight} proposed one of the popular  solution to model-free reinforcement learning (RL). HER facilitates ``sample-efficient'' learning in ``multi-goal'' RL settings using ``binary'' rewards by enabling a robot (an agent) to extract learning opportunities from failed attempts. This is achieved by replaying trajectories with modified goal definitions, allowing the agent to interpret failures as alternative successes. In the context of ``multi-goal'' RL, HER objective is to train policy neural networks capable of achieving a diverse set of goals, say, for example, enabling a robotic manipulator to move an object to various positions on a table top setup. Whereas, ``binary rewards'', in the HER framework, refer to assigning a simple negative or non-negative reward signal to a transition, reflecting either the failure or success of the task as a whole. The formal definition of this reward function $R$ is as follows:
\begin{equation} \label{eq:binary_reward}
\begin{split}
r_t = R(s_t ,a_t , s_{t+1}, g)=\textbf{1}\left[ d(s_{t+1},g) \le \epsilon_{R} \right] -1,
\end{split}
\end{equation}

where, $\textbf{1}$ denote the indicator function, $d$ a metric measure (e.g., the Euclidean distance between the position of the object in $s_{t+1}$ and the goal $g$), and $\epsilon_{R} > 0$ is a fixed threshold. It is very convenient to deploy such a reward signal, as no fine-tuning is needed.

Despite HER notable advantages, it is limited by the absence of a principled replay mechanism; instead it relied on hand-crafted heuristics. Additionally, achieving higher policy accuracy exacerbates the inefficiency of the learning process due to the limitations of sparse non-negative rewards in binary reward functions. Since the development of highly accurate policies is critical for enabling arm robot to perform complex tasks with precision and reliability, particularly in domains such as manufacturing, construction, and healthcare, it is imperative to propose an approach that facilitate more sample-efficient learning of highly accurate multi-goal policies for solving complex robotic-arm tasks. To this end, we propose an novel approach dubbed as ``Next-Future'' that improves the sample-efficiency of the HER based algorithms for training accurate policies without loosing sample-efficiency. \cref{fig:teaser-function} shows an example result using our approach to the robotic arm task (for example: push an object to a desired location). In this paper, we make the following contributions:


\smallskip
\formattedparagraph{Contributions.}

\begin{itemize}
    \item \textbf{Novel Approach:} We present a principled approach to goal relabeling that provide excellent learning signal to the agent. Our approach facilitates more accurate policy learning and improved sample efficiency under stringent performance requirements.

    \item \textbf{Enhanced Value Learning:} By incorporating single-step transition rewards into the multi-goal RL setting, we improve the Q-function learning process, thereby improving overall performance in tasks that demand precision.
    
    \item \textbf{Broad Applicability:} We show that integrating our approach with existing HER-based methods, such as Energy-based Hindsight Experience Prioritization (EBP) \cite{zhao2018energy} and Curriculum-guided Hindsight Experience Replay (CHER) \cite{fang2019curriculum}, leads to further gains, demonstrating the versatility and impact of our method.
    
\end{itemize}

\section{RELATED WORK}\label{sec:rel_work}
\noindent
Sample-efficiency, as defined by \cite{botvinick2019reinforcement}, refers to the quantity of data required by a learning system to achieve a specified performance threshold. This represents a significant challenge in reinforcement learning (RL), which has prompted the development of various methodological innovations. The following sections analyze both model-based and model-free RL approaches that have demonstrated significant improvements in sample efficiency.

\smallskip
\formattedparagraph{\textit{(i)} Model-based RL.} Model-based methods learn a world model that represents the dynamics of the environment and use it to predict future states and rewards, enabling robot planning. The incorporation of a world model enables agents to evaluate potential actions by inferring their long-term consequences, thus enhancing sample-efficiency through predictive planning. Hafner et al. \cite{hafner2019dream} demonstrate efficient solutions to long-horizon tasks through latent imagination, specifically by propagating analytic gradients of learned state values backward through trajectories derived from a learned world model. Moreover, predictive models can be integrated with planning algorithms; Ebert et al. \cite{ebert2018visual} implemented deep predictive models and established that visual model predictive control (MPC) effectively addresses various object manipulation challenges utilizing a learned model.

\smallskip
\formattedparagraph{\textit{(ii)} Model-free RL.} Model-free approaches establish direct mappings between states and actions without explicitly modeling environmental dynamics. These methodologies bifurcate into online and offline paradigms. Online RL involves continuous agent-environment interaction \cite{ghasemi2024comprehensive, ball2023efficient}, whereas offline RL enables agent training on pre-existing datasets, eliminating the resource-intensive online data collection phase. This necessitates maximal utilization of available data resources.

Levine et al. \cite{levine2016end} developed a guided policy search algorithm for fixed datasets that achieves policy learning with significantly reduced interactions compared to traditional online RL methods. Nevertheless, offline RL presents challenges related to distributional shifts between training data and real-world interaction distributions. Consequently, online RL can serve as an effective fine-tuning mechanism to enhance policies initially developed through offline RL pre-training, an approach demonstrated in \cite{julian2020efficient, kalashnikov2018scalable}.

Within this context, the introduction of replay buffers by Lin et al. \cite{lin1992self} represents a significant advancement, enabling RL algorithms to reuse historical data multiple times. In contrast to on-policy methods that utilize each data point only once, the iterative processing of a replay buffer amplifies the learning signal. Building upon this foundation, Schaul et al. \cite{schaul2015prioritized} introduced prioritized experience replay, demonstrating that preferentially selecting transitions with higher temporal difference error (TD-error) within the replay buffer substantially accelerates the training process.

\smallskip
\formattedparagraph{\textit{(iii)} Leveraging Human Expertise.} A methodical approach to enhancing sample efficiency involves the integration of human expert demonstrations. Kalashnikov et al. \cite{kalashnikov2018qt} combined imitation learning with RL to achieve improved sample-efficiency in robotic manipulation tasks, enabling learning from minimal expert demonstrations. The utilization of human expert experiences proves highly effective in refining RL policies. Consequently, initializing a policy through imitation learning before RL refinement has become an established methodology. The aforementioned work by Levine et al. \cite{levine2016end} exemplifies this approach by training neural networks to replicate human expert actions before implementing RL for policy refinement.

\smallskip
\formattedparagraph{\textit{(iv)} Hierarchical and Curriculum Learning.} An alternative approach to improving sample-efficiency is through Hierarchical RL. This methodology addresses the challenge that learning complex robotic tasks from first principles requires extensive trial-and-error exploration. Conversely, acquiring individual sub-tasks that constitute the complete task typically requires fewer interactions. Following this principle, Yang et al. \cite{yang2020multi} proposed a modular learning algorithm that decomposes tasks into constituent sub-tasks and facilitates simultaneous learning across these components. Their research demonstrated enhanced performance in robotic arm manipulation tasks with improved sample-efficiency.

Curriculum learning \cite{narvekar2020curriculum} represents a structured approach to agent training through progressively increasing task difficulty. This methodology has demonstrated significant improvements in data-efficiency. Within this framework, Florensa et al. \cite{florensa2017reverse} implemented reverse curriculum generation, wherein robots learn to reach goal states from initial conditions that progressively increase in distance from the goal. In subsequent research, Florensa et al. \cite{florensa2018automatic} developed automated sub-task generation methods, operating on the principle that mastering sub-tasks facilitates the subsequent acquisition of more complex objectives.

\smallskip
\formattedparagraph{\textit{(v)} Goal-Conditioned Reinforcement Learning.} Researchers have  exploited the divergence between executed environmental behaviors and optimized training behaviors through goal-conditioned RL. Andrychowicz et al. \cite{andrychowicz2017hindsight} implemented a system to replay unsuccessful trajectories with virtual goals, thereby encouraging agents to learn from actual achievements rather than failures. Zhao et al. \cite{zhao2018energy} developed an energy-based experience prioritization framework that samples trajectory transitions with probabilities corresponding to their energies. Fang et al. \cite{fang2019curriculum} established that failed experiences possess differential value across learning phases and proposed a selection strategy based on proximity to target goals and exploration across diverse objectives. These methodologies collectively demonstrate enhanced sample efficiency.

\section{PRELIMINARIES}\label{sec:preliminaries}

\formattedparagraph{Markov Decision Process.} In the  RL framework, an agent interacts with an environment commonly formulated as a Markov Decision Process (MDP), which can be either fully-observed or partially-observed. For clarity and simplicity in subsequent discussions, we focus on fully-observable MDPs. Formally, a fully-observable MDP is defined as a tuple $\left\langle S,A,P,R,\rho_0 \right\rangle$ where $S$ represents the state space containing all possible environmental states, $A$ denotes the complete action space available to the agent, $P: S \times A \to [0,1]$ specifies the state transition probability with $P(s^\prime \mid s,a)$ defining the probability of transitioning to state $s^\prime$ when action $a$ is executed in state $s$. The reward function $R: S \times A \to \mathbb{R}$ quantifies the immediate reward $r_t = R(s_t ,a_t)$ received upon taking action $a_t$ in state $s_t$ at time step $t$, and $\rho_0$ represents the initial state distribution from which the starting state is sampled. Within this mathematical framework, the fundamental objective of reinforcement learning is to discover an optimal policy that maximizes the expected cumulative (often discounted) reward over time.

\smallskip
\formattedparagraph{Reinforcement Learning.} RL operates through a continuous feedback loop between an intelligent agent and its environment. RL is governed by well-defined mathematical principles. At each time step $t$, the agent observes the environment's current state $s_t \in S$, then selects an action $a_t \in A$ using its policy $\pi: S \to A$. This policy---whether deterministic or stochastic---serves as the agent's decision-making mechanism, mapping perceived states to actions. Upon executing $a_t$, the environment transitions to a new state $s_{t+1}$ according to the probabilistic dynamics model $P(s_{t+1} \mid s_t,a_t)$, while simultaneously generating a scalar reward signal $r_t = R(s_t, a_t)$. This cyclical process generates trajectories $\tau = (s_0, a_0, s_1, a_1,...)$, which encapsulate the agent’s entire interaction history within the environment. The core objective in RL is to identify an optimal policy $\pi^*$ that maximizes the expected cumulative discounted return:
\begin{equation}
    J(\pi) = \mathbb{E}_{\tau \sim \pi}{[R(\tau)]},
\end{equation}

where, $R(\tau)$ is the finite-horizon discounted return referring to the sum of all rewards of a fixed-length trajectory each discounted by the discount factor $\gamma$ as in $R(\tau) = \sum_{t=0}^{T} \gamma^{t}r_t$. $\gamma$ balances immediate versus long-term rewards, ensuring convergence in infinite-horizon settings. The optimization problem is formally expressed as:

\begin{equation}
\begin{aligned}
& \displaystyle  \pi^* = \arg \max_{\pi} J(\pi)\\
& \displaystyle \text{where}, \pi^* \text{is the sought-after optimal policy}    
\end{aligned}
\end{equation}

\smallskip
\formattedparagraph{Multi-Goal RL.} We formalize the multi-goal reinforcement learning problem through a Markov Decision Process (MDP) defined by the tuple $\left\langle S,A,P,R,\rho_0, G \right\rangle$, where $G$ denotes the additional goal space. The objective involves learning a unified policy $\pi(s_t, g)$ capable of achieving arbitrary goals $g \in G$, such as manipulating objects to specific target positions or navigating to distinct spatial coordinates.

At each episode initialization, a goal $g$ is sampled from the goal space $G$, and the agent executes actions through its goal-conditioned policy $\pi(s_t, g)$. Following Universal Value Function Approximators (UVFA) principles from \cite{schaul2015universal}, the goal conditioning is achieved by concatenating states $s \in S$ and goals $g \in G$ into extended representations $(s,g)$ such that value function approximators $V(s,g)$ and $Q(s,a)$ can be generalized as $V(s,a,g)$ and $Q(s,a,g)$. This enables generalized value function approximations that operate over the combined state-goal space. The optimization objective remains $\pi^* = \arg \max_{\pi} J(\pi)$, with all functional components now explicitly goal-parameterized---a formulation we designate as multi-goal reinforcement learning.

We can interpret this framework as simultaneously learning a continuum of standard MDPs, each associated with a distinct goal $g \in G$. As illustrated in \cref{fig:value-function-gMDP}, the multi-goal MDP structurally embeds these individual goal-specific MDPs such that their collective goal spaces span $G$. The compositional nature of this architecture enables knowledge transfer across goals while maintaining theoretical consistency with fundamental MDP properties.

\section{OBSERVATION AND METHODOLOGY}\label{sec:background}

\subsection{Observation}
Consider an example of a robot pushing a rectangular prism with continuous state and action spaces; the objective is to move the prism to achieve a desired position and orientation. When trained using binary rewards, standard reinforcement learning (RL) algorithms struggle to learn a goal-conditioned policy in this environment. This difficulty arises due to the low probability of encountering a non-negative reward, which significantly hinders effective learning. To address this issue, Hindsight Experience Replay (HER) \cite{andrychowicz2017hindsight} was introduced as a mechanism to augment training by retrospectively relabeling trajectories with virtual goals that would have resulted in successful outcomes, thereby increasing the frequency of non-negative rewards.

\begin{figure}[t]
    \centering
    \includegraphics[height=3.1cm, keepaspectratio]{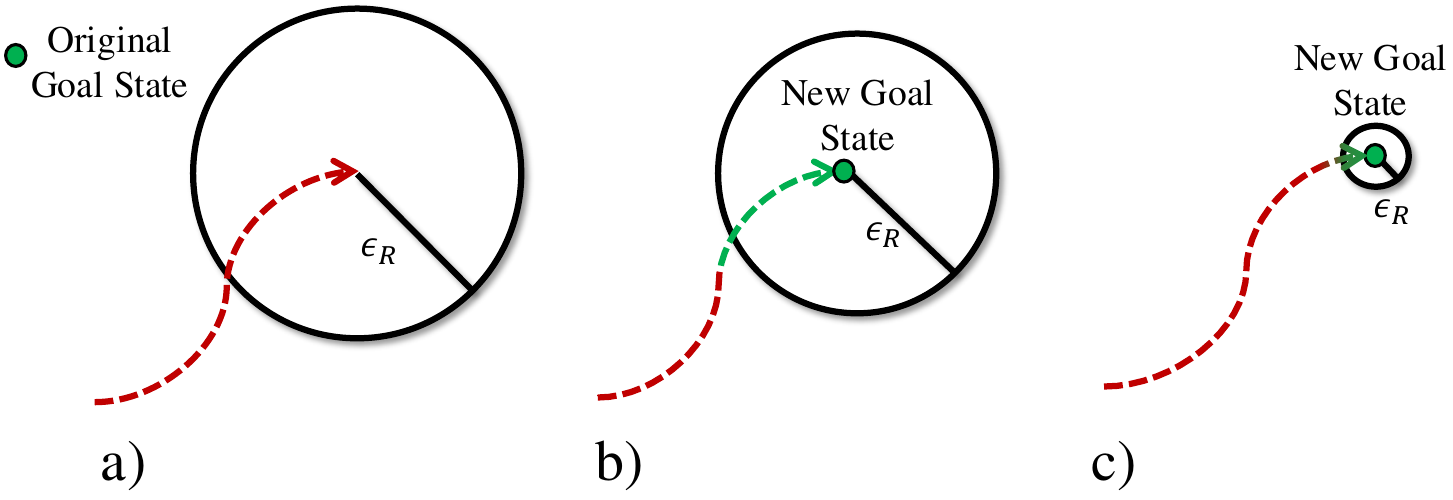}
    \caption{Illustration of goal-conditioned trajectories where transitions are color-coded with their respective binary rewards (red for negative and green for non-negative reward). a) All transitions in the original trajectory receive a negative reward since the goal state is more than $\epsilon_{R}$ away from each state. This makes learning difficult for standard RL algorithms. b) If each transition is replayed where the final state of the environment is considered as the new goal then those nearby it will be associated with a non-negative reward and learning will be facilitated. This corresponds to \emph{Final} strategy of HER work \cite{andrychowicz2017hindsight}. c) Reducing $\epsilon_{R}$ to improve the accuracy of the policy eliminates most of the rewards and HER's performance degrades.}
    \label{fig:trajectory-rewards}
\end{figure}

However, when the success threshold in the binary reward function is tightened, learning becomes substantially more challenging. Our experimental results indicate that the performance of HER also deteriorates under such stringent accuracy requirements, akin to other RL algorithms. Specifically, HER exhibits slower learning progress and requires a significantly large number of trials to achieve comparable success rates. We hypothesize that this degradation is due to the dependency of the HER's effective reward signal on the success threshold. As the threshold becomes more stringent, the number of transitions associated with nonnegative rewards diminishes, reducing the overall effectiveness of HER in facilitating policy learning. An illustration of the same is provided in \cref{fig:trajectory-rewards}.

\begin{figure}[t]
    \centering
    \includegraphics[height=6cm, keepaspectratio]{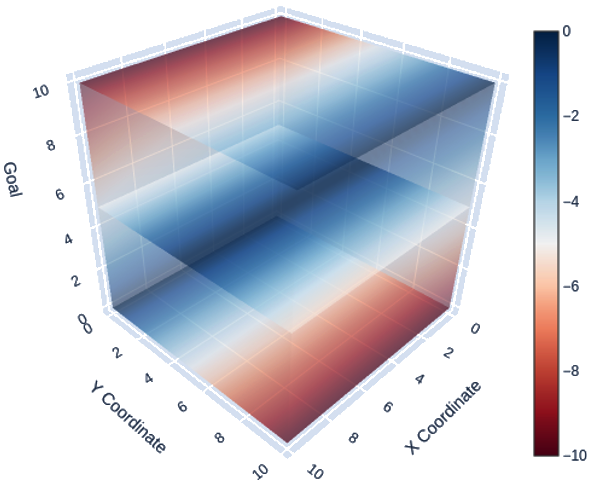}
    \caption{Value function illustration of an MDP for multi-goal RL. Three horizontal slices taken from this multi-goal MDP represent separate standard MDPs with their unique goals. While the state-space is in 2D (x and y coordinates) the goal space is only 1D and matches the y-coordinate of the agent. It is observed visually that goal-conditioning the value function alters the underlying MDP differently across the goal space.}
    \label{fig:value-function-gMDP}
\end{figure}

\subsection{Methodology}
In this work, we introduce an efficient learning based approach for multi-goal RL problem that can solve popular robotic-arm task efficiently. We propose to use the following formulation for  $J(\pi)$ gradient computation for multi-goal RL:
\begin{equation} \label{eq:policy_gradient-1}
\begin{split}
 \nabla_{\theta} J(\theta) = \mathbb{E}_{\pi}\Bigl[\nabla_{\theta} \log \pi(a \mid s, g) Q^{\pi}(s, a, g)\Bigr], ~\text{where,}
\end{split}
\end{equation}
\begin{equation} \label{eq:policy_gradient-2}
\begin{split}
 Q^{\pi}(s, a, g) = \mathbb{E}_{\pi}\Bigl[\sum_{t=0}^{T} \gamma^{t} r(s_t ,a_t , s_{t+1}, g) \Big| s_{0} = s, a_{0} = a\Bigr]
\end{split}
\end{equation}

Here, $\theta$ denotes the parameters of the policy $\pi$. As is known, due to the sparse nature of binary rewards, it is not useful to use \cref{eq:policy_gradient-1} and \cref{eq:policy_gradient-2} to train the model that leads to reliable convergence. Instead, in fact, there needs to be an adequate variability among the observed rewards for it to correctly and reliably compute the gradient.  

HER \cite{andrychowicz2017hindsight} proposes to replay hindsight goals to convert failures into successes and consequently increase the number of non-negative rewards encountered.  HER stores in its replay buffer the original state transitions, e.g. $(s_{k}, g, a_{k}, r_{k}, s_{k+1})$, written in tuple not only with the original goal but also with a subset of other goals, e.g. $(s_{k}, g', a_{k}, r_{k}, s_{k+1})$, for instance the state received at the end of the trajectory $g' = s_{T}$. This way it creates an implicit curriculum without a need to hand-craft it. This simple trick is what speeds up the learning of model-free RL algorithms.
It introduces the following four strategies for the intermediate goal selection step. 1) \emph{Future:} replay with $k$ random states that come after the transition in the same episode, 2) \emph{Final:} replay with the final state from the same episode, 3) \emph{Episode:} replay with $k$ random states from the same episode, and 4) \emph{Random:} replay with $k$ random states from the whole replay buffer.
HER concluded that \emph{Future} with $k = 4$ performs the best, where $k$ determines the ratio of HER transitions to normal replay buffer transitions.

\begin{figure*}
    \centering
    \includegraphics[scale=0.55]{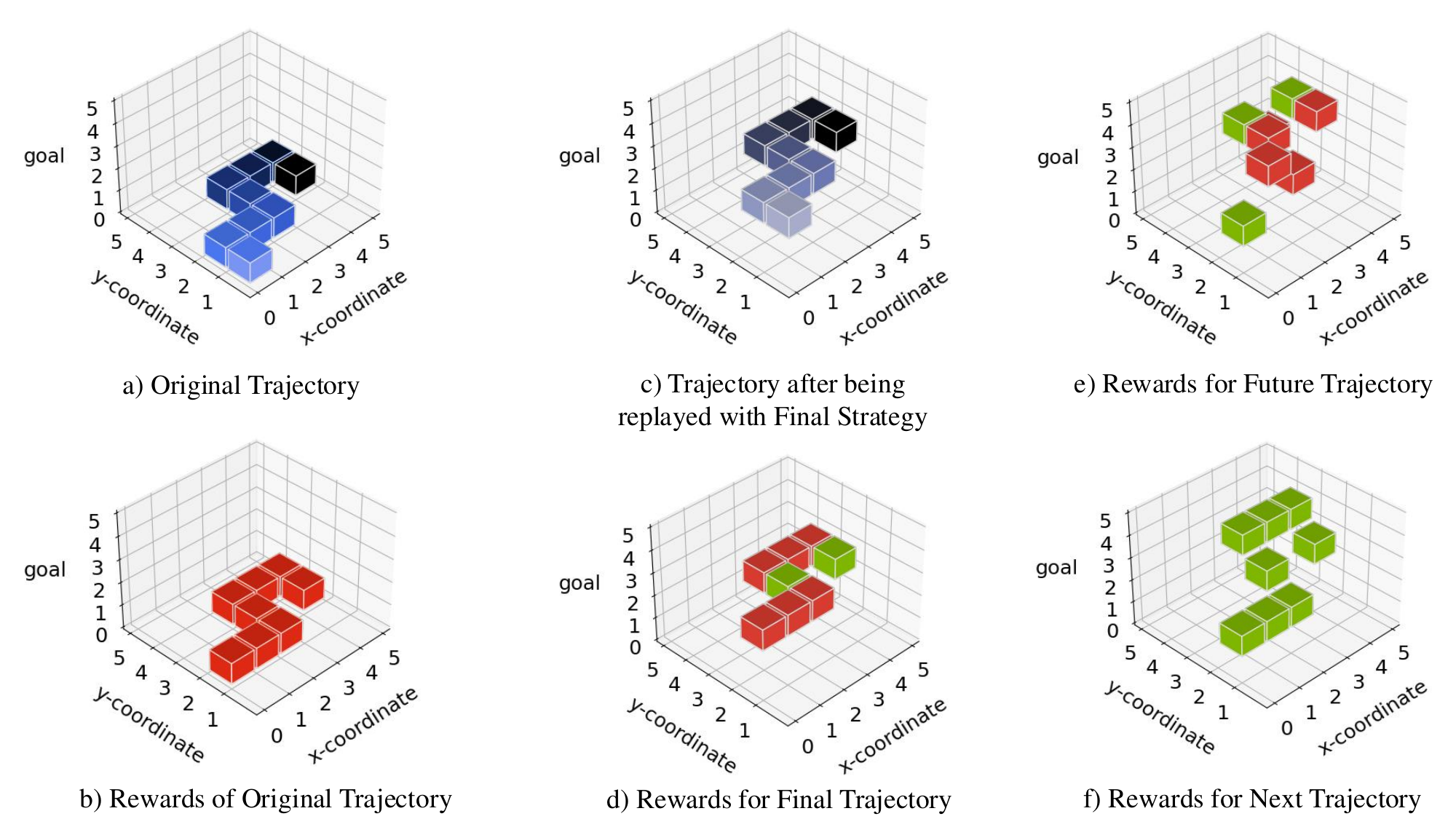}
    \caption{Illustration of state transitions in a multi-goal MDP and their corresponding rewards under different goal selection strategies (red for negative and green for non-negative reward). a) and b): All transitions of the original trajectory receive negative rewards because no transition has achieved the goal of attaining zero y-coordinate. c) and d): Applying \emph{Final} goal selection strategy lifts the original trajectory up in the goal axis to the last achieved state of the episode. Consequently, last transition is awarded non-negatively. An intermediate transition is also given a non-negative reward as it happens to achieve the same goal state after the augmentation. e) and f) strategy randomly moves each state vertically in the goal coordinate and there is no certainty about how the rewards will change. On the other hand, \emph{Next} strategy deterministicly converts each transition to a successful one by altering the goal state as the next achieved state.}
    \label{fig:goal-selection-methods}
\end{figure*}

\smallskip
We propose a simple and effective virtual goal selection technique, named ``Next-Future'', that enables sample-efficient learning under strict policy accuracy requirements. The core idea behind our approach is to guarantee at least one non-negative reward in the augmented transitions by selecting the ``next'' achieved state as the virtual goal for the first of $k$ many trajectory replays. With this, all transitions augmented with ``next'' become successful ones and given a non-negative reward. The remaining $k-1$ virtual goals are then selected uniformly from \emph{Future} achieved states to enable fast distribution of values to non-adjacent states. Despite being simple, this has vital importance since objects in the environment are usually initialized far away from their goal-state and our agents need to execute a long sequence of actions before the task can be fulfilled. Concretely,  we define a ``Next-Future'' approach to generate better training samples:
\begin{enumerate}
    \item \textbf{Next}: Relabel each transition \((s_t, a_t, s_{t+1}, g)\) with \(g_{\text{next}} = s_{t+1}\). This guarantees non‐negative rewards for those transitions.
    \item \textbf{Future}: For the remaining \(k-1\) relabelings, randomly pick \emph{Future} states from the same trajectory to serve as the virtual goal. This helps Q-values propagate to transitions where the goal is farther away.
\end{enumerate}
\cref{fig:goal-selection-methods} demonstrates the effect of our and other goal selection strategies on how they transfer transitions in a multi-goal MDP by using a discretized version of setup shown in \cref{fig:value-function-gMDP} as an example. In its essence, altering the goal state of a transition means changing its place in a multi-goal MDP across the goal dimension. Since we can pick any goal from the goal-space we know that there is at least one goal-state we can augment our transition with, and it will for sure result in a non-negative reward. Our method ``Next-Future'' unlike the HER strategies such as \emph{Final} and \emph{Future} exploits this fact and chooses next achieved state as the new goal. Consequently, all transitions replayed with \textbf{Next} receive non-negative rewards. The remaining $k-1$ virtual goals are then selected randomly from \textbf{Future} achieved states such that Q-values propagate to states with far-away goals, like the ones we expect to encounter when a new episode is initialized and the object is far away from its goal-state.

Since overestimation bias is one of the major hindrances to accurate off-policy learning and since accurate approximation of the Q-function is crucial for sample efficient learning, we adhere to truncated quantile critics (TQC) in our formulation. 

\smallskip
\noindent
\textbf{1) Truncated Quantile Critics (TQC).} Unlike a standard critic $Q^{\pi} (s, a, g)$ that estimates the scalar expected return, we maintain a distributional critic. 
This help mitigate overestimation bias inherent in actor‐critic algorithms. Thus, instead of a single scalar \(Q^\pi\), TQC \cite{kuznetsov2020controlling} learns a distributional critic:
\[
  Z^\pi(s,a,g) 
  \;=\; 
  \bigl[Z^\pi_1(s,a,g), \,Z^\pi_2(s,a,g), \ldots, Z^\pi_N(s,a,g)\bigr],
\]
where each \(Z^\pi_i\) is modeled by a neural network head that outputs multiple quantile points for the return distribution. To reduce the overestimation, we perform truncation. At target update time, TQC gathers all quantile values from the target critics, sorts them, and discards the top \(\tau\)-fraction. Averaging only the \((1-\tau)\)-fraction that remains leads to a more conservative (lower) bias estimate. We define:
\begin{equation}
  \label{eq:tqc_target}
  \operatorname{TQ}(\hat{Z}) \;=\;
  \text{Mean}\Bigl(\text{Sorted}(\{\hat{z}_{i,j}\}) \setminus \text{top } \tau\Bigr),
\end{equation}

where \(\hat{z}_{i,j}\) is the \(j\)-th quantile of the \(i\)-th target critic head, and \(\tau\) is the truncation fraction (e.g., \(0.2\)).

\smallskip
\noindent
\textbf{2) Critique Update.} We maintain a replay buffer \(\mathcal{B}\) that stores transitions (both real and relabeled). For a sampled transition \((s,a,s',g,r)\), the \textit{target} for each quantile head in the critic is
\[
  y \;=\; 
  r 
  \;+\;
  \gamma \,(1 - d)\,
  \operatorname{TQ}\!\Bigl(\hat{Z}^\pi(s', \pi_\theta(s',g), g)\Bigr),
\]
where \(d\) is an indicator for terminal states. Each head performs quantile regression using the Huber quantile loss:
\[
\ell(z^i_j, y) \;=\; \sum_{m=1}^{M}
\rho_{\hat{\tau}_m}\bigl(y_m - z^i_j\bigr),
\]

where \(\hat{\tau}_m\) is the \(m\)-th quantile fraction (e.g., $\frac{2m-1}{2M}$), $y_m$ is the $m^{th}$ sample from the target distribution $y$, and \(\rho\) is the appropriate quantile Huber loss.

\smallskip
\noindent
\textbf{3) Policy Update.} After updating the critics, we update the policy by maximizing the \emph{truncated mixture} Q-value:
\[
Q^\pi_{\mathrm{TQ}}(s,a,g) 
\;=\;
\operatorname{TQ}\!\Bigl(
  \{Z^\pi_1(s,a,g), \ldots, Z^\pi_N(s,a,g)\}
\Bigr).
\]
Hence, the gradient step for the policy becomes:
\begin{equation}\label{eq:policy_update}
\begin{aligned}
  & \nabla_{\theta} J(\theta) \;\approx\;\\
  &   \mathbb{E}_{(s,g)\sim \mathcal{B}}
  \Bigl[
    \nabla_{\theta}\,\pi_\theta(s,g)\,\nabla_a\,
    Q^\pi_{\mathrm{TQ}}(s,a,g)\bigl\vert_{a=\pi_\theta(s,g)}
  \Bigr].
\end{aligned}
\end{equation}

Note that the critic is updated by quantile regression against a truncated target distribution, mitigating overestimation. The policy is updated to maximize the truncated mixture $Q_{TQ}^{\pi}$ which provides a more conservative estimate of $Q^{\pi}$.

Unlike HER \cite{andrychowicz2017hindsight}, it is observed to provide a more accurate approximation of Q-function by alleviating the overestimation bias through three pivotal ideas: distributional representation of critic, truncation of critics prediction, and ensembling of multiple critics. Consequently, this formulation helps us improve the overall stability and sample efficiency of our multi-goal deep RL approach for solving robotic-arm tasks.

\section{EXPERIMENTS}\label{sec:experiments}
We test the performance and robustness of our method by training policies with random seeds for eight challenging robotic arm tasks. Our results show a significant improvement in sample efficiency in seven out of eight tasks while at the same time achieving a higher success rate in six tasks.

\smallskip
\formattedparagraph{Implementation Details.} We use RL Baselines3 Zoo \cite{rl-zoo3} training framework for reinforcement learning. It uses RL algorithm implementations from Stable Baselines3 \cite{stable-baselines3}, which requires Python 3.7+ and PyTorch $\geq$ 1.11. We use a simulation package based on the PyBullet physics engine called panda-gym \cite{gallouedec2021panda}. We use the available hyperparameters for the manipulation tasks considered by RL Baselines3 Zoo \cite{rl-zoo3}. We approximate the policy and the value function using fully connected neural networks with three hidden layers of 512 neurons each. We employed our approach with two critics to overcome bias estimation in Q-learning-based approaches. We train our agents with a learning rate of $ 10^{-3}$, a batch size of $2048$, a replay buffer of length $ 10^{6}$, and a discount factor $\gamma$ of $0.99$. Apart from the last layer, rectified linear unit (ReLU) activation functions are used. In the last layer, hyperbolic trigonometric tangent (tanh) transformation is introduced to squash the action and handle the bounds more correctly.

\smallskip
\formattedparagraph{Baselines and Ablations.}
The baseline for our experiments is the HER method \cite{andrychowicz2017hindsight} which is the state-of-the-art algorithm for training multi-goal agents using a binary reward function. We use it with the \emph{Future} goal selection strategy, as it gives the best results among other replay methods. For ablations, we implemented our formulation with couple of popular methods, namely Energy-based Hindsight Experience Prioritization (EBP) \cite{zhao2018energy} and Curriculum-guided Hindsight Experience Replay (CHER) \cite{fang2019curriculum}, showing the suitability of our ``Next-Future'' approach.

\begin{figure*}[t]
    \centering
    \includegraphics[scale=0.56]{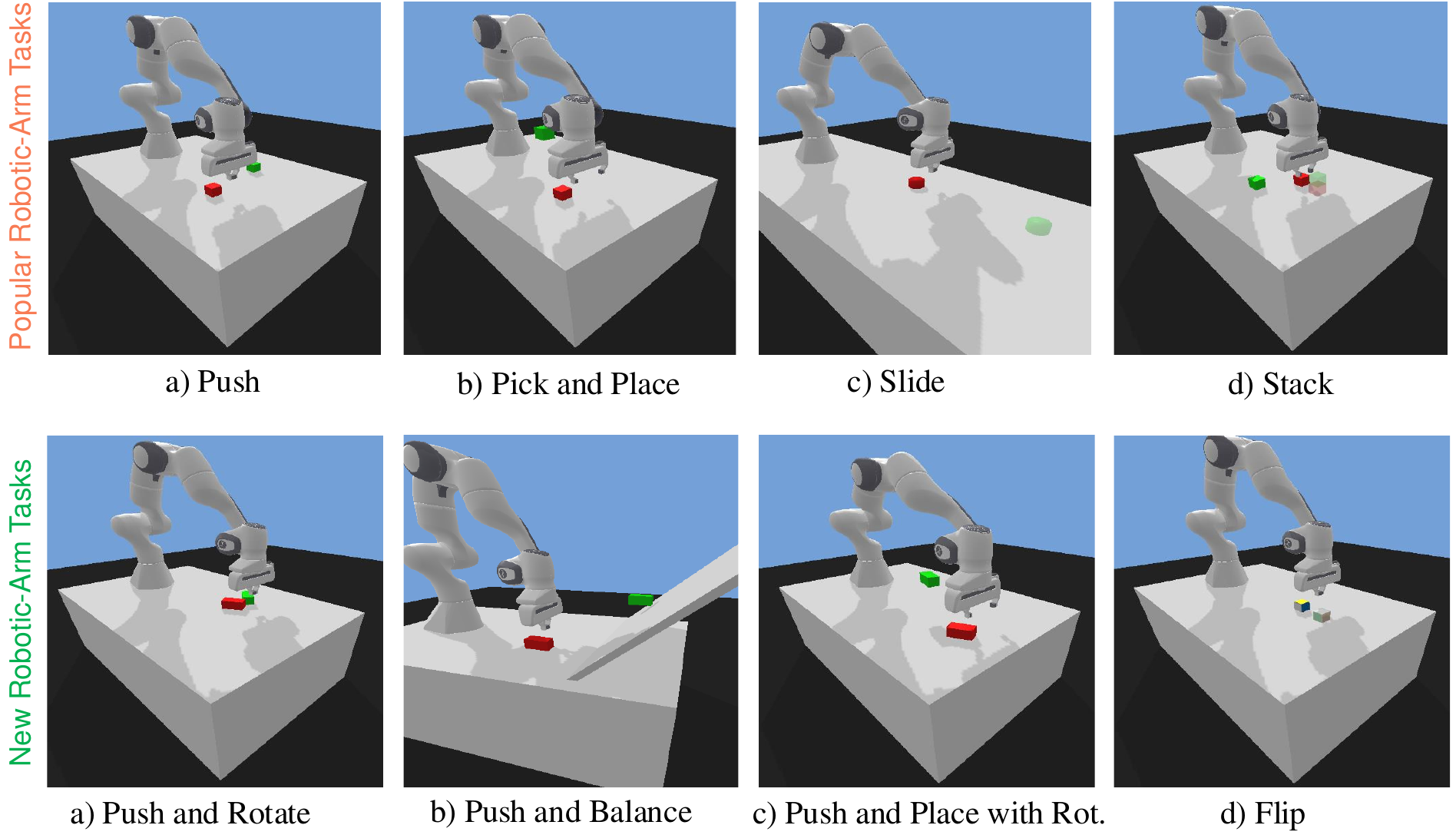}
\caption{Experimental setup in simulation for the popular and newly introduced robotic-arm tasks, respectively.}
\label{fig:tasks}
\end{figure*}

\subsection{Robot-Arm Tasks}
Below, we detail different robotic arm tasks we considered for this work. For benchmarking purposes, we split them into two groups, namely the a) Popular tasks and b) Newer tasks. All tasks we study in this work involve moving an object to a desired position and, in specific tasks, also to a desired orientation. The results of a visual simulation of each task are shown in \cref{fig:tasks}. For results using the real-world robotic arm, refer to the supplementary video provided with this paper. 

\smallskip
\noindent
The popular tasks that we tested are:
\begin{itemize}
    \item {\fontfamily{cmtt}\selectfont{Push.}} This task involves bringing a cube to a desired position on the table. Gripper fingers are kept fully closed at all times to prevent learning to grasp the object and to promote pushing it instead. 
    \item {\fontfamily{cmtt}\selectfont{Pickup-and-Place.}} Similar to the pushing task, a cube needs to be brought to a desired position but, this time, the goal position can be above the table (e.g., in the free space such as air). For the agent to be able to grasp the object, it is allowed to move its fingers.
    \item {\fontfamily{cmtt}\selectfont{Slide.}} In this task a puck needs to be slided over a long slippery table top such that it stops at a desired location, which is outside of robot's reach, solely due to friction.  
    \item {\fontfamily{cmtt}\selectfont{Stack.}} It requires stacking two cubes at a desired position with the green cube over the red one. For quantitative evaluations, we used stacking of two objects, yet our approach showed successful results on stacking three objects at a given locations as well (refer supp. video).
\end{itemize}

\smallskip
\noindent
The newer tasks that we introduced and tested are:
\begin{itemize}
    \item {\fontfamily{cmtt}\selectfont{Push-and-Rotate.}} The pushing task is extended such that the goal condition also includes a desired orientation of the object. 
    \item {\fontfamily{cmtt}\selectfont{Push-and-Balance.}} The pushing task is extended such that a plane with a $45^{\circ}$ tilt angle is placed at the far end of the table and the goal position is selected across this tilted plane. The agent needs to learn to balance the cube at height using the tilted plane as a support. 
    \item {\fontfamily{cmtt}\selectfont{Pickup-and-Place with Rotation.}} The pick-and-place is extended with additional goal criteria to place the cube in the correct orientation. Here, the robot is allowed to rotate its end-effector around its z-axis.
    \item  {\fontfamily{cmtt}\selectfont{Flip.}} It requires flipping a cube and bringing it to a desired position with the correct orientation. Here, the cube can be initialized with any side facing the top view.
\end{itemize}

\smallskip
\noindent
\formattedparagraph{Reward Function.} We use binary rewards and differentiate between success and failure of the task, as defined in \cref{eq:binary_reward}. A task is considered fulfilled if the Euclidean distance between the object's position and the goal position is less than a certain threshold. For tasks with additional orientation criteria, the success condition further involves a threshold in radians between the object's attained and desired orientation.

\smallskip
\noindent
\formattedparagraph{Observation Space.} All tasks possess certain common measurements as well as some task-specific observations. Measurements of the end-effector absolute position and linear velocities are common for all environments. If it is allowed to control the orientation of the gripper around its z-axis, then its measurement in radians is included in the observation space too. Additionally, for each object present in the scene, the absolute position, orientation, and linear and angular velocities are introduced to the observation vector. Lastly, the distance between the fingers is also observed in tasks where the gripper fingers are free to move. Note that all orientations are represented as quaternions.

\smallskip
\formattedparagraph{Action Space.} In all environments, the robotic arm can be moved freely in the 3D space. Therefore, all tasks include the desired relative\footnote{In this paragraph \emph{relative} means relative to the current state.} gripper movement in their action space. If the fingers are actuated then there is an action unit corresponding to the desired variation of the gripper opening. Moreover, for tasks where the end-effector can rotate around its z-axis there is also a corresponding action that represents the desired relative rotation in radians.

\subsection{Evaluations}

\formattedparagraph{\textit{(i)} Success Rate.} 
A common way to evaluate robotic‐arm tasks with sparse‐reward reinforcement learning is success rate. It is the fraction of episodes in which the goal is achieved at least once. Formally, consider $N$ evaluation episodes under policy $\pi$, each with an indicator $\mathbb{I}\text{(goal achieved in episode $i$)}$. The success rate is then defined as:
\begin{equation}
    \text{SuccessRate}(\pi) = \sum_{i=1}^{N}\mathbb{I}[\text{goal achieved in episode $i$}].
\end{equation}

\begin{figure*}[t]
    \centering
    \includegraphics[scale=0.25]{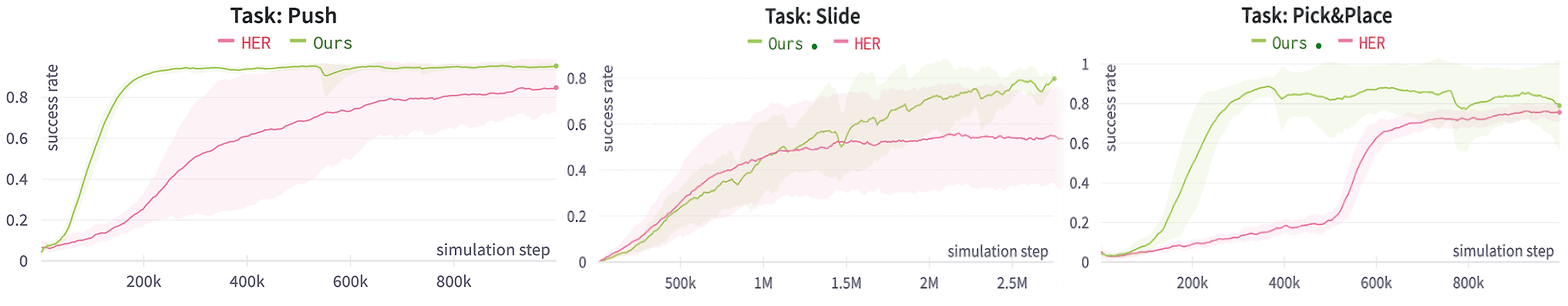}
\caption{Success rate on different robotic-arm tasks compared to HER \cite{andrychowicz2017hindsight}.}
\label{fig:success-rate-curve}
\end{figure*}

\cref{fig:success-rate-curve} shows the train time success rate curve over simulation step for a few well-known robotic arm task. \cref{tb:max_sr} lists the maximum success rates obtained with the HER algorithm when using ours ``Next-Future'' and HER's \emph{Future} goal selection strategies. Our method achieves an equal or higher maximum success rate in seven out of eight manipulation tasks. This demonstrates the effectiveness of our method in training accurate policies. 

\begin{table}[ht]
    \centering
    \begin{tabular}[t]{lcc}
    \toprule
    Tasks & Next-Future (Ours) & Future  \\ \midrule
      Push    &        \textbf{1.00 $\pm$ 0.004}     &          \textbf{1.00 $\pm$ 0.003}     \\
      Pick\&Place   &             \textbf{0.95 $\pm$ 0.03}    &      0.82 $\pm$ 0.29      \\
      Slide    &              0.52 $\pm$ 0.17     &          \textbf{0.62 $\pm$ 0.05 }     \\
      Stack   &            \textbf{0.93 $\pm$ 0.02}      &          0.90 $\pm$ 0.01      \\ 
      Push\&Rotate    &          \textbf{0.93 $\pm$ 0.02}    &          0.90 $\pm$ 0.03       \\
      Push\&Balance   &              \textbf{0.84 $\pm$ 0.06}      &          0.81 $\pm$ 0.03        \\
      Pick\&Place w/ Rot.    &          \textbf{0.74 $\pm$ 0.39}       &          0.64 $\pm$ 0.44       \\
      Flip   &             \textbf{0.81 $\pm$ 0.04}       &          0.64 $\pm$ 0.05       \\
      \bottomrule
    \end{tabular}
    \caption{Maximum success rate comparison with the HER algorithm. Entries with bold characters represent the better-performing approach.}
    \label{tb:max_sr}
\end{table}

\noindent
\formattedparagraph{\textit{(ii)} Sample-efficiency.}  To measure the effect of our method on the sample efficiency, we compare the number of data-points it requires to reach a certain level of success rate for the first time. The results regarding the same is reported in \cref{tb:se}. 

In seven out of eight tasks our method significantly outperforms the \emph{Future} strategy. The sliding task is again the only task where \emph{Future} has a better sample-efficiency. Overall, our method gives a superior sample-efficiency performance thanks to its fast Q-value estimation achieved through greedily rewarding transitions by adjusting their goals definitions.

\begin{table}[ht]
    \centering
    \begin{tabular}[t]{lcc}
    \toprule
    Tasks & Next-Future (Ours) & Future  \\ \midrule
      Push    &        $\boldsymbol{0.22 \cdot 10^{6}}$     &          $0.39 \cdot 10^{6} $     \\
      Pick\&Place   &       $ \boldsymbol{0.16 \cdot 10^{6}}$  & $0.44 \cdot 10^{6}$       \\
      Slide    &              $0.40 \cdot 10^{6}$     &  $\boldsymbol{0.32 \cdot 10^{6}}$   \\
      Stack   &             $\boldsymbol{0.53 \cdot 10^{6}}$         &          $0.99 \cdot 10^{6}$       \\ 
      Push\&Rotate    &          $\boldsymbol{0.32 \cdot 10^{6}} $    &          $ 0.49 \cdot 10^{6}$       \\
      Push\&Balance   &              $ \boldsymbol{0.26 \cdot 10^{6}} $       &          $0.47 \cdot 10^{6}$       \\
      Pick\&Place w/ Rot.    &          $ \boldsymbol{0.45 \cdot 10^{6}} $       &         $  0.8 \cdot 10^{6} $       \\
      Flip   &            $\boldsymbol{0.64 \cdot 10^{6}}$     &        $1.00 \cdot 10^{6}$   \\
      \bottomrule
    \end{tabular}
    \caption{Sample-efficiency comparison with HER algorithm. Entries with bold characters represent the better-performing approach.}
    \label{tb:se}
\end{table}
\FloatBarrier

\smallskip
\noindent
\underline{\textit{Reason for a lower performance in the sliding task.}} It can be observed from the \cref{tb:max_sr} and \cref{tb:se} that for the sliding task, HER's \cite{andrychowicz2017hindsight} \emph{Future} strategy outperforms our ``Next-Future'' method. This is because mastering this task requires learning the long-horizon consequences of attaining high object speeds due to previously taken actions. In the sliding task, once the object is released and out of the end-effector's reach, it moves solely due to its accumulated velocity. At this stage, the agent has no means to interact with the object. Since our method boosts the sample efficiency by learning single-step transitions, it does not bring much contribution to learning tasks such as sliding. On the contrary, as seen in our experiments, it may even confuse and deteriorate the learning process.

\begin{figure*}[t]
    \centering
    \includegraphics[scale=0.52]{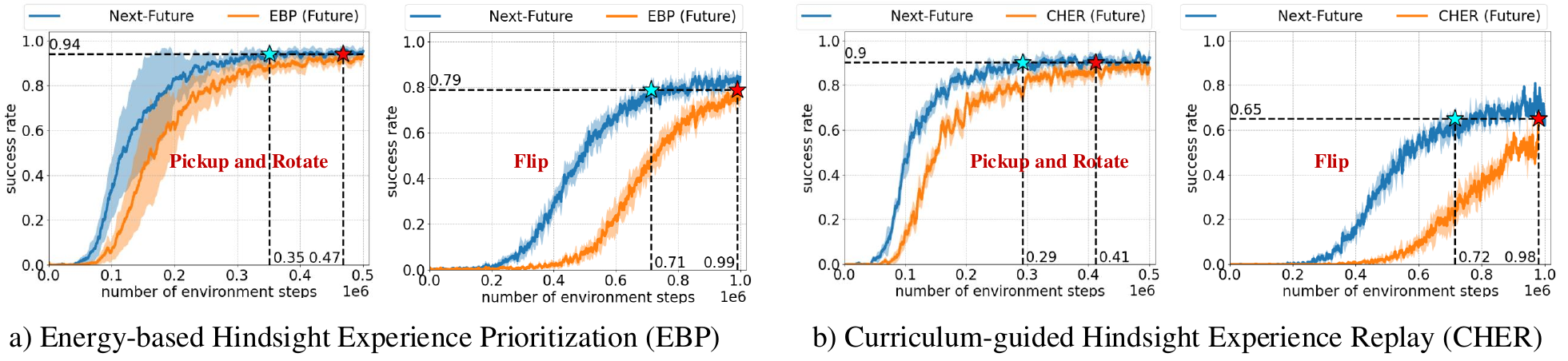}
\caption{a)-b) Learning curves comparison of our Next-Future with Future goal selection methods using EBP \cite{zhao2018energy} and CHER \cite{fang2019curriculum} for Pickup and Rotate and Flip Task, respectively. The results are averaged across 10 random seeds and shaded areas represent one standard deviation. Here, we compare replay strategies based on two metrics: the maximum success rate and the sample efficiency. Clearly, our Next-Future provides a better replay strategy for learning accurate policies efficiently.}
\label{fig:ablations-comparison}
\end{figure*}

\smallskip
\formattedparagraph{\textit{(iii)} Ablation Study.} We experimentally evaluate how the ``Next-Future'' goal selection strategy affects the sample efficiency with other popular Hindsight Experience Replay approaches. To this end, we picked Energy-based Hindsight Experience Prioritization (EBP) \cite{zhao2018energy} and Curriculum-guided Hindsight Experience Replay (CHER) \cite{fang2019curriculum} for our ablations. We introduced our formulation to their pipeline with our own goal selection strategy to conduct this ablation study. \cref{tb:max_4gw4gsr_newer} and \cref{tb:max_sgwer_newer} provides the success rate and sample-efficiency using EBP \cite{zhao2018energy} algorithm, while \cref{tb:max_4gw4gsr_newer-cher} and \cref{tb:max_sgwer_newer-cher}  provides the success rate and sample-efficiency using CHER \cite{fang2019curriculum} algorithm compared to ours. The statistical results follow the same trend as before. Additionally, we provide the learning curves on a couple of robotic-arm tasks in \cref{fig:ablations-comparison}, demonstrating benefits of our approach. Overall, these ablations shows that our approach can provide better sample efficiency and success rate in comparison to Hindsight Experience Replay approaches in seven out of eight manipulation tasks. The only task where our method slightly falls behind is the sliding task.

\begin{table}
    \centering
    \begin{tabular}[t]{lcc}
    \toprule
    Tasks & Next-Future (Ours) & Future  \\ \midrule
      Push    &        \textbf{1.00 $\pm$ 0.00}     &          \textbf{1.00 $\pm$ 0.00}     \\
      Pick\&Place   &             \textbf{0.99 $\pm$ 0.006}    &      0.99 $\pm$ 0.007      \\
      Slide    &              0.52 $\pm$ 0.11     &          \textbf{0.59 $\pm$ 0.07}     \\
      Stack   &            \textbf{0.92 $\pm$ 0.03}      &          0.90 $\pm$ 0.04      \\ 
      Push\&Rotate    &          \textbf{0.95 $\pm$ 0.02}    &          0.94 $\pm$ 0.02       \\
      Push\&Balance   &              \textbf{0.83 $\pm$ 0.04}      &          0.77 $\pm$ 0.08       \\
      Pick\&Place w/ Rot.    &          \textbf{0.95 $\pm$ 0.03}       &          0.93 $\pm$ 0.03       \\
      Flip   &             \textbf{0.84$ \pm$ 0.03}       &          0.78 $\pm$ 0.02       \\
      \bottomrule
    \end{tabular}
    \caption{Maximum success rate comparison with the EBP \cite{zhao2018energy}. Entries with bold characters represent the better-performing approach.}
    \label{tb:max_4gw4gsr_newer}
\end{table}

\begin{table}
    \centering
    \begin{tabular}[t]{lcc}
    \toprule
    Tasks & Next-Future (Ours) & Future  \\ \midrule
      Push    &       $\boldsymbol{0.23 \cdot 10^{6}}$    &          $0.36 \cdot 10^{6} $    \\
      Pick\&Place   &       $ \boldsymbol{0.22 \cdot 10^{6}}$   & $0.50 \cdot 10^{6}$       \\
      Slide    &              $0.76 \cdot 10^{6}$     &  $ \boldsymbol{0.38 \cdot 10^{6}}$   \\
      Stack   &             $\boldsymbol{0.61 \cdot 10^{6}}$         &          $0.90 \cdot 10^{6}$       \\ 
      Push\&Rotate    &          $\boldsymbol{0.35 \cdot 10^{6}} $    &         $ 0.47 \cdot 10^{6}$      \\
      Push\&Balance   &              $ \boldsymbol{0.22 \cdot 10^{6}} $     &          $0.28 \cdot 10^{6}$      \\
      Pick\&Place w/ Rot.    &           $ \boldsymbol{0.63 \cdot 10^{6}} $      &         $  0.86 \cdot 10^{6} $      \\
      Flip   &            $\boldsymbol{0.71 \cdot 10^{6}}$      &         $0.99 \cdot 10^{6}$   \\
      \bottomrule
    \end{tabular}
    \caption{Sample-efficiency comparison with the EBP \cite{zhao2018energy}. Entries with bold characters represent the better-performing approach.}
    \label{tb:max_sgwer_newer}
\end{table}

\begin{table}
    \centering
    \begin{tabular}[h]{lcc}
    \toprule
    Tasks & Next-Future (Ours) & Future  \\ \midrule
      Push    &        \textbf{1.00 $\pm$ 0.004}     &          \textbf{1.00 $\pm$ 0.004}     \\
      Push\&Rotate    &          \textbf{0.95 $\pm$ 0.01}    &          $0.90 \pm 0.01$       \\
      Push\&Balance   &              \textbf{0.65 $\pm$ 0.33}      &          $0.57 \pm 0.32$        \\
      Flip   &             \textbf{0.81 $\pm$ 0.04}       &          $0.64 \pm 0.05$       \\
      \bottomrule
    \end{tabular}
    \caption{Maximum success rate comparison with the CHER \cite{fang2019curriculum}. Entries with bold characters represent the better-performing approach.}
    \label{tb:max_4gw4gsr_newer-cher}
\end{table}

\begin{table}
    \centering
    \begin{tabular}[h]{lcc}
    \toprule
    Tasks & Next-Future (Ours) & Future  \\ \midrule
      Push    &        $\boldsymbol{0.22 \cdot 10^{6}}$      &          $ 0.27 \cdot 10^{6}$   \\
      Push\&Rotate    &          $\boldsymbol{0.29 \cdot 10^{6}}$    &          $0.41 \cdot 10^{6} $      \\
      Push\&Balance   &             $ \boldsymbol{0.22 \cdot 10^{6}}$       &          $0.50 \cdot 10^{6}$      \\
      Flip   &            $\boldsymbol{0.72 \cdot 10^{6}}$    &        $0.98 \cdot 10^{6}$   \\
      \bottomrule
    \end{tabular}
    \caption{Sample-efficiency comparison with CHER \cite{fang2019curriculum}. Entries with bold characters represent the better-performing approach.}
    \label{tb:max_sgwer_newer-cher}
\end{table}

\smallskip
\formattedparagraph{\textit{(iv)} Analysis.} Accurately learning the Q‐values is critical to training a high‐performing policy. 
\begin{figure} 
    \centering
    \includegraphics[scale=0.24]{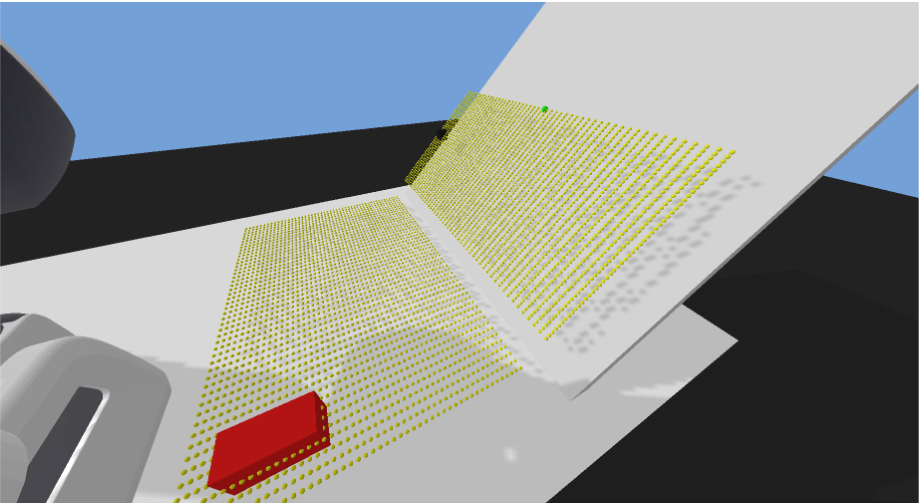}
    \caption{Q-function is queried while the end-effector holds the object still at the visualized yellow marks. The goal is selected as moving the object to the green mark.}
    \label{fig:query-points}
\end{figure}
And therefore, we assessed how our approach influences the learned Q-function. In particular, we evaluated the Q function with a zero action vector while the end-effector balances the object with the yellow markers shown in \cref{fig:query-points}, in order to transfer it to the green marker located at the center of the final row of yellow markers in the tilted plane. Because the underlying reward function is binary, we expect higher Q-values for yellow markers positioned nearer to the goal, where reaching the goal state is comparably more feasible. Conversely, markers farther from the goal should yield lower Q-values. Note that all value estimates remain non-positive, in keeping with the reward signal being strictly $-1$ or $0$.

\begin{figure}[t]
    \centering
    \includegraphics[height=3.8cm, keepaspectratio]{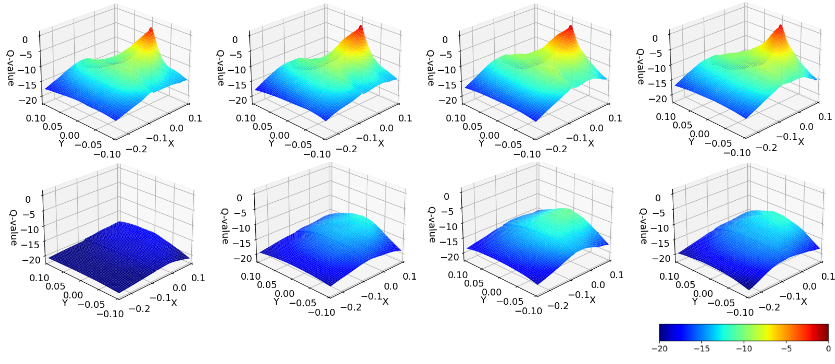}
\caption{Q-values learned using the goal selection strategy ``Next-Future'' and \emph{Future}, respectively visualized at every 30000 training steps.}
\label{fig:q-values}
\end{figure}

\cref{fig:q-values} illustrates the estimated Q-values at the queried object locations with ``Next-Future'' (top row) and \emph{Future} (bottom row), respectively, and HER algorithm is used with the goal selection strategies. We observed that our goal selection strategy can estimate the Q-values correctly and quickly. In contrast, the ``Future'' strategy can only estimate some part of the Q-values at a much slower pace. The \emph{Future} method especially struggles to estimate the Q-values corresponding to the query points on the tilted plane. We argue that this is due to the fact that these states are not well-represented in the replay buffer because they are not experienced much during the early stages of the training, as moving the object across the tilted plane requires developing a sense of balance first. On the other hand, our ``Next-Future'' strategy can cope with this difficulty---thanks to rewarding single-step transitions that enable value propagation across successive states in trajectories.

\section{DISCUSSION}
As alluded to above, our method outperforms other baselines on all tasks except the sliding task. Nevertheless, we anticipate its effectiveness on the sliding task as well, provided the goal state includes the object's velocity. In that case, incorporating velocity‐based rewards into single‐step transitions would eliminate rewarding transitions where the object exhibits high velocity and cannot maintain position. In this setting, we expect our approach to further improve sample efficiency for the sliding task. We aim to extend our work in this direction in the future utilizing object(s) instance \cite{zurbrugg2024icgnet} or semantic  \cite{menini2021real} \cite{haghighi2023neural} priors. Furthermore, in the future, we aspire to solve robotic arm tasks for soft and deforming objects, utilizing non-rigid structure-from-motion method output priors from \cite{kumar2022organic} \cite{kumar2020dense} as input to our policy design.

\section{CONCLUSION}
This paper introduced a novel approach for sample-efficient deep reinforcement learning (DRL) tailored to robotic-arm tasks. The proposed framework demonstrates significant improvements in sample efficiency relative to the current state-of-the-art algorithms. We performed rigorous experimentation across diverse benchmark tasks to show the usefulness of our approach. Critically, policies trained via our approach exhibit robust performance when deployed in real-world robotic systems without requiring additional fine-tuning. Empirical results suggest that this work advances the field by providing a systematic framework for reducing the sample complexity of DRL for robotic arm tasks while maintaining functional generality. We hope these contributions will stimulate further investigation into foundational aspects of sample efficiency and generalization in DRL for robotic arm applications.






\section*{ACKNOWLEDGMENT}
The authors thank VIS group leader at ETH Z\"urich for a few helpful discussion.

{\small
\bibliographystyle{IEEEtran}
\bibliography{IEEEfull,root}

\begin{thebibliography}{10}
\providecommand{\url}[1]{#1}
\csname url@samestyle\endcsname
\providecommand{\newblock}{\relax}
\providecommand{\bibinfo}[2]{#2}
\providecommand{\BIBentrySTDinterwordspacing}{\spaceskip=0pt\relax}
\providecommand{\BIBentryALTinterwordstretchfactor}{4}
\providecommand{\BIBentryALTinterwordspacing}{\spaceskip=\fontdimen2\font plus
\BIBentryALTinterwordstretchfactor\fontdimen3\font minus \fontdimen4\font\relax}
\providecommand{\BIBforeignlanguage}[2]{{%
\expandafter\ifx\csname l@#1\endcsname\relax
\typeout{** WARNING: IEEEtran.bst: No hyphenation pattern has been}%
\typeout{** loaded for the language `#1'. Using the pattern for}%
\typeout{** the default language instead.}%
\else
\language=\csname l@#1\endcsname
\fi
#2}}
\providecommand{\BIBdecl}{\relax}
\BIBdecl

\bibitem{murray2017mathematical}
R.~M. Murray, Z.~Li, and S.~S. Sastry, \emph{A mathematical introduction to robotic manipulation}.\hskip 1em plus 0.5em minus 0.4em\relax CRC press, 2017.

\bibitem{levine2016end}
S.~Levine, C.~Finn, T.~Darrell, and P.~Abbeel, ``End-to-end training of deep visuomotor policies,'' \emph{The Journal of Machine Learning Research}, vol.~17, no.~1, pp. 1334--1373, 2016.

\bibitem{yu2020meta}
T.~Yu, D.~Quillen, Z.~He, R.~Julian, K.~Hausman, C.~Finn, and S.~Levine, ``Meta-world: A benchmark and evaluation for multi-task and meta reinforcement learning,'' in \emph{Conference on robot learning}.\hskip 1em plus 0.5em minus 0.4em\relax PMLR, 2020, pp. 1094--1100.

\bibitem{henderson2018deep}
P.~Henderson, R.~Islam, P.~Bachman, J.~Pineau, D.~Precup, and D.~Meger, ``Deep reinforcement learning that matters,'' in \emph{Proceedings of the AAAI conference on artificial intelligence}, vol.~32, no.~1, 2018.

\bibitem{yu2018towards}
Y.~Yu, ``Towards sample efficient reinforcement learning.'' in \emph{IJCAI}, 2018, pp. 5739--5743.

\bibitem{andrychowicz2017hindsight}
M.~Andrychowicz, F.~Wolski, A.~Ray, J.~Schneider, R.~Fong, P.~Welinder, B.~McGrew, J.~Tobin, O.~Pieter~Abbeel, and W.~Zaremba, ``Hindsight experience replay,'' \emph{Advances in neural information processing systems}, vol.~30, 2017.

\bibitem{zhao2018energy}
R.~Zhao and V.~Tresp, ``Energy-based hindsight experience prioritization,'' in \emph{Conference on Robot Learning}.\hskip 1em plus 0.5em minus 0.4em\relax PMLR, 2018, pp. 113--122.

\bibitem{fang2019curriculum}
M.~Fang, T.~Zhou, Y.~Du, L.~Han, and Z.~Zhang, ``Curriculum-guided hindsight experience replay,'' \emph{Advances in neural information processing systems}, vol.~32, 2019.

\bibitem{botvinick2019reinforcement}
M.~Botvinick, S.~Ritter, J.~X. Wang, Z.~Kurth-Nelson, C.~Blundell, and D.~Hassabis, ``Reinforcement learning, fast and slow,'' \emph{Trends in cognitive sciences}, vol.~23, no.~5, pp. 408--422, 2019.

\bibitem{hafner2019dream}
D.~Hafner, T.~Lillicrap, J.~Ba, and M.~Norouzi, ``Dream to control: Learning behaviors by latent imagination,'' \emph{arXiv preprint arXiv:1912.01603}, 2019.

\bibitem{ebert2018visual}
F.~Ebert, C.~Finn, S.~Dasari, A.~Xie, A.~Lee, and S.~Levine, ``Visual foresight: Model-based deep reinforcement learning for vision-based robotic control,'' \emph{arXiv preprint arXiv:1812.00568}, 2018.

\bibitem{ghasemi2024comprehensive}
M.~Ghasemi, A.~H. Mousavi, and D.~Ebrahimi, ``Comprehensive survey of reinforcement learning: From algorithms to practical challenges,'' \emph{arXiv preprint arXiv:2411.18892}, 2024.

\bibitem{ball2023efficient}
P.~J. Ball, L.~Smith, I.~Kostrikov, and S.~Levine, ``Efficient online reinforcement learning with offline data,'' in \emph{International Conference on Machine Learning}.\hskip 1em plus 0.5em minus 0.4em\relax PMLR, 2023, pp. 1577--1594.

\bibitem{julian2020efficient}
R.~Julian, B.~Swanson, G.~S. Sukhatme, S.~Levine, C.~Finn, and K.~Hausman, ``Efficient adaptation for end-to-end vision-based robotic manipulation,'' in \emph{4th Lifelong Machine Learning Workshop at ICML 2020}, 2020.

\bibitem{kalashnikov2018scalable}
D.~Kalashnikov, A.~Irpan, P.~Pastor, J.~Ibarz, A.~Herzog, E.~Jang, D.~Quillen, E.~Holly, M.~Kalakrishnan, V.~Vanhoucke \emph{et~al.}, ``Scalable deep reinforcement learning for vision-based robotic manipulation,'' in \emph{Conference on Robot Learning}.\hskip 1em plus 0.5em minus 0.4em\relax PMLR, 2018, pp. 651--673.

\bibitem{lin1992self}
L.-J. Lin, ``Self-improving reactive agents based on reinforcement learning, planning and teaching,'' \emph{Machine learning}, vol.~8, pp. 293--321, 1992.

\bibitem{schaul2015prioritized}
T.~Schaul, J.~Quan, I.~Antonoglou, and D.~Silver, ``Prioritized experience replay,'' \emph{arXiv preprint arXiv:1511.05952}, 2015.

\bibitem{kalashnikov2018qt}
D.~Kalashnikov, A.~Irpan, P.~Pastor, J.~Ibarz, A.~Herzog, E.~Jang, D.~Quillen, E.~Holly, M.~Kalakrishnan, V.~Vanhoucke \emph{et~al.}, ``Qt-opt: Scalable deep reinforcement learning for vision-based robotic manipulation,'' \emph{arXiv preprint arXiv:1806.10293}, 2018.

\bibitem{yang2020multi}
R.~Yang, H.~Xu, Y.~Wu, and X.~Wang, ``Multi-task reinforcement learning with soft modularization,'' \emph{Advances in Neural Information Processing Systems}, vol.~33, pp. 4767--4777, 2020.

\bibitem{narvekar2020curriculum}
S.~Narvekar, B.~Peng, M.~Leonetti, J.~Sinapov, M.~E. Taylor, and P.~Stone, ``Curriculum learning for reinforcement learning domains: A framework and survey,'' \emph{Journal of Machine Learning Research}, vol.~21, no. 181, pp. 1--50, 2020.

\bibitem{florensa2017reverse}
C.~Florensa, D.~Held, M.~Wulfmeier, M.~Zhang, and P.~Abbeel, ``Reverse curriculum generation for reinforcement learning,'' in \emph{Conference on robot learning}.\hskip 1em plus 0.5em minus 0.4em\relax PMLR, 2017, pp. 482--495.

\bibitem{florensa2018automatic}
C.~Florensa, D.~Held, X.~Geng, and P.~Abbeel, ``Automatic goal generation for reinforcement learning agents,'' in \emph{International conference on machine learning}.\hskip 1em plus 0.5em minus 0.4em\relax PMLR, 2018, pp. 1515--1528.

\bibitem{schaul2015universal}
T.~Schaul, D.~Horgan, K.~Gregor, and D.~Silver, ``Universal value function approximators,'' in \emph{International conference on machine learning}.\hskip 1em plus 0.5em minus 0.4em\relax PMLR, 2015, pp. 1312--1320.

\bibitem{kuznetsov2020controlling}
A.~Kuznetsov, P.~Shvechikov, A.~Grishin, and D.~Vetrov, ``Controlling overestimation bias with truncated mixture of continuous distributional quantile critics,'' in \emph{International Conference on Machine Learning}.\hskip 1em plus 0.5em minus 0.4em\relax PMLR, 2020, pp. 5556--5566.

\bibitem{rl-zoo3}
A.~Raffin, ``Rl baselines3 zoo,'' \url{https://github.com/DLR-RM/rl-baselines3-zoo}, 2020.

\bibitem{stable-baselines3}
A.~Raffin, A.~Hill, A.~Gleave, A.~Kanervisto, M.~Ernestus, and N.~Dormann, ``Stable-baselines3: Reliable reinforcement learning implementations,'' \emph{Journal of Machine Learning Research}, vol.~22, no. 268, pp. 1--8, 2021.

\bibitem{gallouedec2021panda}
Q.~Gallou{\'e}dec, N.~Cazin, E.~Dellandr{\'e}a, and L.~Chen, ``panda-gym: Open-source goal-conditioned environments for robotic learning,'' \emph{4th Robot Learning Workshop: Self-Supervised and Lifelong Learning at NeurIPS}, 2021.

\bibitem{zurbrugg2024icgnet}
R.~Zurbr{\"u}gg, Y.~Liu, F.~Engelmann, S.~Kumar, M.~Hutter, V.~Patil, and F.~Yu, ``Icgnet: a unified approach for instance-centric grasping,'' in \emph{2024 IEEE International Conference on Robotics and Automation (ICRA)}.\hskip 1em plus 0.5em minus 0.4em\relax IEEE, 2024, pp. 4140--4146.

\bibitem{menini2021real}
D.~Menini, S.~Kumar, M.~R. Oswald, E.~Sandstr{\"o}m, C.~Sminchisescu, and L.~Van~Gool, ``A real-time online learning framework for joint 3d reconstruction and semantic segmentation of indoor scenes,'' \emph{IEEE Robotics and Automation Letters}, vol.~7, no.~2, pp. 1332--1339, 2021.

\bibitem{haghighi2023neural}
Y.~Haghighi, S.~Kumar, J.-P. Thiran, and L.~Van~Gool, ``Neural implicit dense semantic slam,'' \emph{arXiv preprint arXiv:2304.14560}, 2023.

\bibitem{kumar2022organic}
S.~Kumar and L.~Van~Gool, ``Organic priors in non-rigid structure from motion,'' in \emph{European Conference on Computer Vision}.\hskip 1em plus 0.5em minus 0.4em\relax Springer, 2022, pp. 71--88.

\bibitem{kumar2020dense}
S.~Kumar, L.~Van~Gool, C.~E. de~Oliveira, A.~Cherian, Y.~Dai, and H.~Li, ``Dense non-rigid structure from motion: A manifold viewpoint,'' \emph{arXiv preprint arXiv:2006.09197}, 2020.

\end{thebibliography}
}

\end{document}